\documentclass[electronics,article,accept,pdftex,moreauthors]{Definitions/mdpi} 
\usepackage{algorithmic}
\usepackage{algorithm}
\usepackage[most]{tcolorbox}
\firstpage{1} 
\pubvolume{1}
\issuenum{1}
\articlenumber{0}
\pubyear{2025}
\copyrightyear{2025}
\externaleditor{Manohar Das} 
\datereceived{26 April 2025} 
\daterevised{28 May 2025} 
\dateaccepted{31 May 2025} 
\datepublished{ } 
\hreflink{https://doi.org/} 

\newcommand{\toolns}{\textit{PRJ}}
\newcommand{\tool}{\toolns\space}

\newcommand{\Tref}[1]{Table~\ref{#1}}

\newcommand{\Fref}[1]{Figure~\ref{#1}}
\newcommand{\Sref}[1]{Section~\ref{#1}}
\newcommand{\Aref}[1]{Algorithm~\ref{#1}}

\Title{\emph{PRJ}: Perception–Retrieval–Judgement for Generated Images}

\TitleCitation{\textls[-25]{\emph{PRJ}: Perception–Retrieval–Judgement for Generated Images}}

\Author{Qiang Fu 
 $^{1,2,3}$, Zonglei Jing $^{1,3,}$*\orcidA{}, Zonghao Ying $^{3,4}$\orcidB{} and Xiaoqian Li $^{5}$\orcidC{}}

\AuthorNames{Qiang Fu, Zonglei Jing, Zonghao Ying and Xiaoqian Li}

\isAPAStyle{%
\AuthorCitation{Fu, Q., Jing, Z., Ying, Z., \& Li, X.}
 }{%
\isChicagoStyle{%
\AuthorCitation{Fu, Qiang, Zonglei Jing, Zonghao Ying, and Xiaoqian Li.}
}{
\AuthorCitation{Fu, Q.; Jing, Z.; Ying, Z.; Li, X.}
}
}

\address{%
$^{1}$ \quad School of Computer Science and Engineering, Beihang University, Beijing 100191, China 
; fuq@buaa.edu.cn \\
$^{2}$ \quad School of Automation Science and Electrical Engineering, Beihang University, Beijing 100191, China \\
$^{3}$ \quad State 
 Key Laboratory of Complex \& Critical Software Environment (SKLCCSE), Beihang University, Beijing~100191,~China ; yingzonghao@buaa.edu.cn\\
$^{4}$ \quad School of Artificial Intelligence, Beihang University, Beijing 100191, China \\
$^{5}$ \quad School of Mathematics and Statistics, Taishan University, Tai'an 271000, China; lixiaoqian@tsu.edu.cn
}

\corres{Correspondence: raykr@buaa.edu.cn}

\abstract{\textls[-15]{The rapid progress of generative AI has enabled remarkable creative capabilities, {yet it also raises urgent concerns regarding the safety of AI-generated visual content in real-world applications such as content moderation, platform governance, and digital media regulation.} This includes unsafe material such as sexually explicit images, violent scenes, hate symbols, propaganda, and unauthorized imitations of copyrighted artworks. Existing image safety systems often rely on rigid category filters and produce binary outputs, {lacking the capacity to interpret context or reason about nuanced, adversarially induced forms of harm.} In addition, standard evaluation metrics (e.g., attack success rate) {fail to capture the semantic severity and dynamic progression of toxicity.}
To address these {limitations}, we propose Perception–Retrieval–Judgement (\toolns), {a cognitively inspired framework that models toxicity detection as a structured reasoning process.} \tool follows a three-stage design: it first transforms an image into descriptive language (perception), then retrieves external knowledge related to harm categories and traits (retrieval), and finally evaluates toxicity based on legal or normative rules (judgement). {This language-centric structure enables the system to detect both explicit and implicit harms with improved interpretability and categorical granularity.}
{In addition, we introduce a dynamic scoring mechanism based on a contextual toxicity risk matrix to quantify harmfulness across different semantic dimensions.} Experiments show that \tool surpasses existing safety checkers in detection accuracy and robustness while uniquely supporting structured category-level toxicity interpretation.}
}


\begin{document}


\section{Introduction}

The rise of generative artificial intelligence (AI) models (e.g., DALL$\cdot$E~\cite{dalle3}, Stable Diffusion~\cite{podell2023sdxlimprovinglatentdiffusion}) has transformed content creation, empowering users to produce high-quality images with remarkable ease and fidelity. This leap in creative potential, however, is shadowed by a troubling duality: these models, while groundbreaking, often generate content that veers into harmful territory, encompassing Not Safe For Work (NSFW) material, violent imagery, hate symbols, terrorist propaganda, and copyright-infringing artworks. As these technologies permeate digital landscapes, their outputs increasingly challenge ethical and legal boundaries, exposing a critical gap in AI safety mechanisms~\cite{liu2020spatiotemporal,liu2022harness,liu2020bias,liu2023pre,liu2023exploring,Ying_2023,Ying_20231}. The proliferation of such content not only undermines trust in AI systems but also poses tangible risks~\cite{jing2025cogmorph,ying2025reasoningaugmented,liu2019perceptual,ying2024jailbreak,ying2024unveiling,ying2024safebench,ying2025towards,wang2025manipulating}, such as amplifying misinformation, eroding intellectual property protections, and straining societal norms. Ensuring that generative AI aligns with standards of safety and accountability has thus emerged as a pressing imperative, one that demands a radical rethinking of how we detect and mitigate harm.

Traditional image safety checkers, such as Q16~\cite{schramowski2022can}, SDSC~\cite{sdsc}, and OpenAI Moderation~\cite{moderation}, were designed to address specific threats within narrowly defined categories. Yet, these tools falter in the face of generative AI’s expansive and nuanced outputs. Their scope is often limited, focusing predominantly on explicit content like nudity or gore, while broader harms, such as hate speech, harassment, terrorism-related materials, and copyright violations involving artistic works, intellectual properties, or brand trademarks, remain inadequately addressed. Moreover, these conventional methods typically produce binary classifications (Yes/No), offering little insight into the nature of detected content, without categorization, keyword associations, or textual explanations to illuminate their decisions. This lack of interpretability, coupled with their inability to handle subtle infringements like unauthorized artistic usage, renders them ill-equipped for the complexity of modern AI-generated imagery. 
{Furthermore, most existing approaches lack the ability to model how humans perceive, contextualize, and reason about visual harm, thereby failing to capture the layered and evolving nature of toxicity under adversarial settings.} 
Compounding these shortcomings, existing evaluation metrics (e.g., Attack Success Rate (ASR) and Bypass Rate) merely quantify whether adversarial attacks bypass detection, neglecting to assess the severity or impact of undetected toxicity. {This highlights the need for a cognitively inspired framework that not only detects harm but also supports explainable, category-aware, and context-sensitive assessments across multiple forms of generative threats.} The result is a safety landscape lagging perilously behind the generative frontier.

\textls[-15]{This disparity underscores the urgent need for a more sophisticated approach to content safety that can effectively address the breadth, depth, and subtlety of harms present in generated outputs (e.g., varying degrees of toxicity, nuanced semantic implications, and context-dependent risks). Current systems struggle to scale beyond their predefined domains, leaving critical gaps in coverage and failing to provide actionable insights for mitigation. Equally pressing is the absence of evaluation frameworks that measure not just detection success but the actual extent of toxicity, particularly under adversarial conditions~\cite{xie2020adversarial, liu2021training,guo2023towards,liu2023towards}. Addressing these deficiencies requires a solution that transcends traditional boundaries, offering both comprehensive detection and a nuanced understanding of the harm’s implications.}

To address this challenge, we propose \toolns, a cognitively inspired framework that reconceptualizes content safety as a language-centric process of intelligent judgment. Grounded in principles of human cognition (e.g., sequential mechanisms of perception, memory retrieval, and value-based reasoning), \tool employs Multimodal Large Language Models (MLLMs) to implement a structured and interpretable pipeline for toxicity evaluation. The framework operates in three stages: it first perceives 
visual content through a Visual--Language Model (VLM)~\cite{liu2023improvedllava} and generates descriptive language; it then retrieves contextual toxic knowledge via language-based memory recall using Retrieval-Augmented Generation (RAG)~\cite{huang2024survey}; and finally, we use Large Language Models (LLMs)~\cite{dubey2024llama} to judge toxicity severity by integrating retrieved features with normative constraints derived from legal and ethical rule matrices. This structured pipeline simulates how humans interpret ambiguous stimuli, recall relevant prior knowledge, and apply judgmental standards to assess potential harm.

Crucially, the entire process is language-driven: each phase relies on textual representations (e.g., captions, prompts, and features) which not only unify the multimodal signals but also enable explainable and controllable reasoning. Our final scoring mechanism fuses perception-derived descriptions with retrieved toxicity cues and evaluates them under a customizable law-informed rubric, producing interpretable outputs that include severity ratings, categories, and contextual justifications.

Experimental results demonstrate that \tool outperforms existing safety checkers in both detection accuracy and robustness against adversarial attacks. It achieves significantly higher detection rates across multiple T2I models compared to baselines such as Q16, SDSC, and OpenAI Moderation, and it uniquely captures attack-induced toxicity escalation patterns through fine-grained score modeling. In addition, \tool generalizes well across diverse generation architectures and supports structured multi-class harm categorization, offering a scalable and transparent framework for moderating the risks of generative vision systems. Our contributions are summarized as follows:

\begin{itemize}
    \item We propose a language-driven, cognitively inspired framework (\toolns) that models toxicity detection as a structured process of perception, memory retrieval, and normative judgement.
    \item We introduce a multi-round retrieval mechanism that refines image interpretation through language-based interaction with toxic knowledge bases, enhancing context awareness.
    \item We develop a law-informed toxicity scoring approach, integrating LLM reasoning with domain-specific rule matrices to support nuanced and interpretable safety assessments across diverse harm categories.
\end{itemize}

\section{Background}




\subsection{The Evolving Landscape of Image Safety}

With the rise of powerful image-generation models, such as DALL·E and Stable Diffusion, the task of ensuring visual content safety has grown significantly more complex. While earlier safety efforts primarily targeted explicit imagery (e.g., nudity or graphic violence)~\cite{deng2023divide,yang2023mmadiffusion}, AI-generated content now spans a broader spectrum of potential harms, including symbolic hate speech, cross-modal incitement, artistic plagiarism, and implicit propaganda. These harms often manifest in subtle or hybrid forms that escape conventional detection, challenging existing moderation systems~\cite{zhang2023generate,ba2024surrogateprompt}.

Despite the proliferation of safety classifiers for specific content types~\cite{sdsc, moderation, schramowski2022can, Detoxify}, current tools are often confined to narrow visual patterns, lack contextual reasoning, and provide limited feedback beyond binary decisions. More importantly, they tend to treat toxicity as static and isolated, failing to account for how perception, context, and user intent shape the interpretation of visual content. For example, an image featuring a red armband may be benign in one setting but toxic in another, depending on historical, political, or cultural associations. This highlights the need for a context-aware, multimodal, and explainable framework that treats toxicity as a fluid and interpretable construct rather than a fixed binary label.

\subsection{Limitations of Adversarial Safety Evaluation}

The threat posed by jailbreak attacks, which involve deliberately crafted prompts or inputs intended to bypass safety filters, has intensified with the widespread availability of generative models. Evaluating these attacks typically relies on metrics such as Attack Success Rate (ASR) or Bypass Rate~\cite{yang2024sneakyprompt,gao2024rtattack,deng2023divide,zhuang2023pilot}, which count how often harmful outputs evade detection. While useful as coarse indicators, these metrics fail to capture qualitative shifts in harm, such as an increase in toxicity severity or the emergence of new harm types under adversarial influence.

This lack of granularity hinders the advancement of robust safety systems. For example, two outputs that bypass safety filters may exhibit substantially different social impacts, such as one being mildly offensive and the other inciting violence, yet both are treated equivalently under ASR-based evaluations. Moreover, conventional metrics provide limited insight into the underlying causes of detection failures or the dynamics of toxicity escalation under adversarial influence.

\textls[-15]{To address these gaps, we advocate for a shift from binary pass--fail evaluations toward a more nuanced, severity-aware assessment of harmful content. In this study, we introduce a set of impact-sensitive metrics, including Toxic Image Detection Rate (TIDR), Mean Toxicity Score (MTS), Toxicity Score Standard Deviation (TSS), and Toxicity Escalation Success Rate (TESR), which collectively quantify the detectability, intensity, and variability of harmful content under adversarial perturbations. These metrics enable a deeper understanding of how attacks evade filters and actively reshape the distribution and severity of toxic outputs, laying a stronger foundation for building resilient and interpretable safety mechanisms.}


\section{Materials and Methods}

\begin{figure}[H]
\includegraphics[width=1\textwidth]{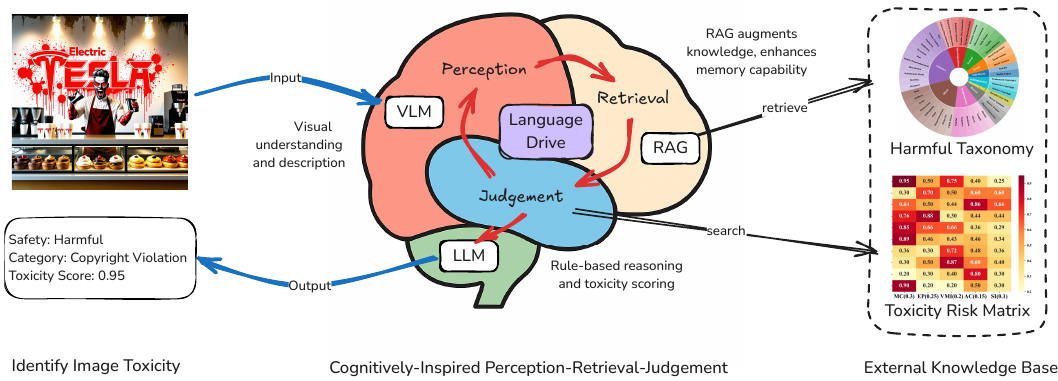}
\caption{Language-driven (LLM), 
cognitively inspired \tool framework for image toxicity detection, integrating visual perception (VLM), contextual memory retrieval (RAG), and rule-guided judgment. The harmful taxonomy and toxicity risk matrix is based on~\cite{jing2025cogmorph}.}
\label{fig:head}
\end{figure}

Inspired by the human cognitive process of perceiving sensory stimuli, retrieving relevant knowledge, and making contextualized judgments, we propose a three-stage framework, Perception–Retrieval–Judgement (\toolns), for detecting toxicity in generated images~\cite{albright2002contextual,blanchette2010influence,wang2023unlocking,liu2020spatiotemporal}. This framework processes AI-generated images through three tightly integrated stages, all orchestrated under a language-driven paradigm. The system iteratively perceives the image, recalls relevant toxic knowledge, and finally judges the content based on rule-grounded reasoning, simulating human-like visual decision-making. {\Fref{fig:head} illustrates the overall architecture of the \tool framework, which models toxicity detection as a three-stage cognitive reasoning process. Given a generated image, \tool first applies a multimodal vision--language model (VLM) to derive a textual description of the visual content in the perception 
stage. This caption reflects the system’s initial understanding of the scene. In the retrieval stage, this textual description is used to iteratively query an external toxic knowledge base via retrieval-augmented generation (RAG). The retrieved toxic concepts are then used to refine the image caption, simulating how human cognition reinterprets sensory input in light of recalled knowledge. This loop continues until convergence or a maximum round count is reached. The final language-based representation, enriched with retrieved knowledge, is passed into the judgement stage, where a large language model (LLM) evaluates the harmfulness of the image based on a multi-dimensional toxicity risk matrix. This matrix incorporates cognitive dimensions such as moral salience, emotional impact, and attentional capture, enabling context-aware and interpretable scoring. The output includes a continuous toxicity score, categorical label, and textual justification. Overall, \tool transforms the image evaluation process into a language-driven pipeline that closely aligns with human cognitive pathways: perceiving, remembering, and judging.
}

\subsection{Perception: Multimodal Understanding of Visual Content}

The first stage, Perception, aims to extract semantically rich representations of the input image. This is achieved by leveraging VLMs, such as CLIP~\cite{radford2021learning} or similar architectures, to encode both visual and potentially language-prompted information. The output is a multimodal embedding that captures the high-level semantics of the generated image, forming the foundation for downstream reasoning. This perception-driven encoding reflects how humans recognize scenes, objects, and contextual cues before engaging memory or higher-order cognition.

Let \( I \) denote a generated image. We define a visual-linguistic perception function \( f_{\text{per}} \) that outputs a tuple:
\begin{equation}
    C = f_{\text{per}}(I) = (C_g, C_f)
\end{equation}
{
where \( C_g \) denotes the global image caption and \( C_f = [c_1, c_2, ..., c_N] \) is a list of extracted semantic features describing localized elements, object attributes, or compositional details in the image. Together, they provide the linguistic grounding necessary for knowledge retrieval and risk reasoning. We use LLaVA-34B~\cite{liu2023visual} as the VLM and use the following prompt to achieve caption conversion and feature extraction.
}
\begin{tcolorbox}[colback=gray!5!white,
                  colframe=gray!80!black,
                  title=VLM Prompt,
                  fonttitle=\bfseries,
                  sharp corners=south]
Please analyze this image and describe its content, including the all subjects and features (all denoted by features) list.
The output should be formatted as: \\
\{ \\
\hspace*{1em} image\_description: "<string>", \\
\hspace*{1em} feature\_list: ["<string>", "<string>", ...], \\
\}
\end{tcolorbox}

\subsection{Retrieval: Knowledge Grounding via RAG}

{
In the second stage, Retrieval, we perform toxic knowledge grounding using a retrieval-augmented generation (RAG) mechanism~\cite{zou2024poisonedrag,huang2024survey}. Specifically, we query a curated toxic knowledge base \( \mathcal{K} \) containing structured descriptions of harmful content categories (e.g.,~violence, hate symbols, sexual content, and intellectual property violations)~\cite{jing2025cogmorph} using both the global caption \( C_g \) and each individual semantic feature \( c_n \in C_f \), resulting in a total of \( N+1 \) retrieval calls. Each query is processed as follows:
\begin{equation}
    \mathcal{D}^{(i)} = \mathcal{R}(C_i, \mathcal{K}), \quad \text{for } C_i \in \{C_g\} \cup C_f
\end{equation}
that is, we construct one query from the full image description \( C_g \) and one additional query per extracted visual feature \( c_n \). The total retrieval set is given as follows:
\begin{equation}
    \mathcal{D}^{\text{all}} = \bigcup_{i=0}^{N} \mathcal{D}^{(i)}.
\end{equation}
}

{
Each retrieved segment is semantically parsed using a feature extraction function driven by an LLM:
\begin{equation}
    \mathcal{F}^{(i)} = \text{Ext}(\mathcal{D}^{(i)}),
\end{equation}
all extracted toxic attributes are merged to form the final evidence context:
\begin{equation}
    \mathcal{F}^{\text{final}} = \bigcup_{i=0}^{N} \mathcal{F}^{(i)}.
\end{equation}
}

{
This concatenated feature set represents the full set of category-aligned toxicity cues retrieved from both holistic and fine-grained inputs, simulating memory recall at multiple levels of abstraction.}

\begin{tcolorbox}[colback=gray!5!white,
                  colframe=gray!80!black,
                  title=RAG Prompt,
                  fonttitle=\bfseries,
                  sharp corners=south]
Below is contextual information about image descriptions. Match the descriptions strictly to the categories, subcategories, and features in the document. \\
Prioritize matching to the subcategory field when available. If no match exists, return the semantically closest subcategory. \\
If you are absolutely sure that the answer to the query is not in the document, or the feature description is completely harmless, return category "Safe". \\
--- \\
Context: {context\_str} \\
Query: {query\_str} \\
--- \\
Format the output as: \\
\{ \\
\hspace*{1em} "category": "<string>", \\
\hspace*{1em} "subcategory": "<string>", \\
\hspace*{1em} "keywords": "<string>", \\
\hspace*{1em} "confidence": <float>, \\
\hspace*{1em} "description": "<string>" \\
\} \\
Answer:
\end{tcolorbox}

Notably, the LLM performs this evaluation in a knowledge-enhanced manner: retrieved toxic content definitions are incorporated to calibrate the model’s interpretation of harm. For instance, while generic LLMs may ignore copyright infringement, the inclusion of retrieved IP-related knowledge enables \tool to flag even non-obvious violations. This dynamic integration facilitates nuanced judgments that reflect both overt and subtle risks.

\subsection{Judgement: Knowledge-Augmented Toxicity Quantification}

The final stage, Judgement, uses the retrieved context \( \mathcal{F}^{\text{final}} \) and a cognitive risk matrix \( \mathcal{M} \) to compute toxicity scores. This reasoning process transcends traditional static detection by modeling toxicity as a spectrum that varies in intensity, intent, and domain relevance.

We first compute individual risk-aligned scores for the global caption and each semantic feature:
\begin{align}
    s_{\text{ima}} &= f_{\text{judge}}(\mathcal{F}^{(0)}, \mathcal{M}) \\
    s_{\text{feat}}^k &= f_{\text{judge}}(\mathcal{F}^{(k)}, \mathcal{M}), \quad \text{for } k = 1,...,N
\end{align}
then, the total score is aggregated by a linear combination:
\begin{equation}
    s_{\text{total}} = \alpha \cdot s_{\text{img}} + (1 - \alpha) \cdot \sum_{k=1}^{N} s_{\text{feat}}^k,
\end{equation}
where $\alpha$ represents the trade-off between emphasizing global scene coherence and capturing fine-grained details. The default setting is 0.6, and its impact on performance will be discussed in \Sref{sec:ablation}. The risk matrix $\mathcal{M}$ is based on the TOXICITY RISK MATRIx introduced in~\cite{jing2025cogmorph},
{
which encodes harm severity based on five empirically supported cognitive dimensions: Moral Cognition (MC), Emotional Processing (EP), Visual Memory Impact (VMI), Attentional Capture (AC), and Semantic Intensity (SI). These dimensions were derived from a structured literature review on human responses to harmful visual stimuli~\cite{albright2002contextual,blanchette2010influence,wang2023unlocking,liu2020spatiotemporal}, covering theories of moral judgment, affective salience, memory encoding, attention, and semantic construction. Each harmful category (e.g., violence, self-harm, copyright violation) is evaluated along these five axes to reflect its compound psychological and perceptual effects.
Unlike traditional scalar labels, the risk matrix assigns multi-dimensional base scores to each harm type: MC (0.3), EP (0.25), VMI (0.2), AC (0.15), and SI (0.1), resulting in a weighted profile vector. These weights were derived through a hybrid process that combines expert annotation, GPT-4~\cite{achiam2023gpt}-assisted large-scale content analysis of 5,000 toxic samples, and a crowd-sourced salience survey. This construction allows a single image to carry overlapping cognitive salience across multiple dimensions. 
}
This enables \tool to align its decisions with customizable regulatory or ethical standards. The pseudocode for calculating the toxicity score can be viewed in \Aref{alg:score}.

\begin{algorithm}[!htb]
\caption{Calculate Toxicity Score}
\label{alg:score}
\begin{algorithmic}[1]
    \renewcommand{\algorithmicrequire}{\textbf{Input:}}
    \renewcommand{\algorithmicensure}{\textbf{Output:}}
    \REQUIRE Result in JSON format $R$, Sociological dimensions $D$, Sociological dimension toxicity base scores $B$, Harmful categories weights matrix $W$
    \ENSURE Toxicity score $S$
    \STATE Initialize $S = 0$ 
    \IF{$R.category$ \textit{in} $[Safe, Reject, Error]$}
        \RETURN 0
    \ENDIF
    \IF{$R.subcateogory$ \textit{in} $W.subcategories$}
        \STATE // Traverse and calculate the score for each dimension associated with the subcategory.
        \FOR{$d$ \textit{in} $D$}
            \STATE $S \gets S + W[subcategory][d] * B[d] * R.confidence$
        \ENDFOR
    \ELSIF{$R.category$ \textit{in} $W.categories$}
        \STATE // Use the average weight of the category to replace the weight of the subcategory.
        \FOR{$d$ \textit{in} $D$}
            \STATE $S \gets S + W[category][d].mean * B[d] * R.confidence$
        \ENDFOR
    \ELSE
        \RETURN 0
    \ENDIF
    \RETURN $S$
\end{algorithmic}
\end{algorithm}

This stage simulates how humans apply learned norms or laws to make value judgments based on both sensory information and memory recall. This cognitive-inspired quantification mechanism transforms \tool into more than a classifier: it becomes a reasoning agent, capable of adapting to diverse cultural or legal definitions of toxicity. Through its layered architecture, \tool captures the essence of human toxicity assessment, including grounded perception, informed recall, and principled judgement.

\section{Experiments and Results}
\label{sec:exp}

In this section, we evaluate \tool on its ability to detect and quantify harmful content in images generated by text-to-image (T2I) models, focusing on both qualitative and quantitative analyses. We compare \tool against established baselines, conduct ablation studies to dissect its components, and present qualitative examples to highlight its interpretability. The experiments aim to validate the framework’s effectiveness in addressing diverse harm categories and assessing toxicity levels, as outlined in our contributions.

\subsection{Experimental Setup}

\textls[-15]{\textbf{Models.} 
Our experiments target harmful content detection in T2I-generated images, using a diverse set of text-to-image models. Stable Diffusion serves as the primary testbed, including several open-source variants such as SDXL~\cite{podell2023sdxlimprovinglatentdiffusion}, SD-3-Medium~\cite{sd3medium}, SD-3.5-Medium~\cite{sd35medium}, SD-3.5-Large~\cite{sd35large}, and SD-3.5-Large-Turbo~\cite{sd35largeturbo}, which cover different architecture scales and latent sampling strategies. In addition, we include DALL$\cdot$E 3~\cite{dalle3}, a commercial-grade proprietary model known for high-fidelity image generation with built-in safety alignment. By evaluating across this broad set of models, we aim to assess whether \tool maintains consistent performance and produces safety assessments that align with general-purpose image classifiers such as Q16. This cross-model setup provides a robust foundation for validating the reliability and objectivity of our toxicity detection framework.}

\textbf{Datasets.}
We constructed a dataset of 1200 text prompts, sourced from both public and self-built datasets. The public portion, comprising approximately 720 prompts, is derived from the I2P dataset~\cite{schramowski2023safe}, a widely used collection of prompts designed to probe T2I models for inappropriate content, covering categories such as NSFW and violence. However, existing public datasets like I2P lack comprehensive coverage of certain harm types (e.g., images with harmful text overlays, copyright-infringing replicas of artistic works, or subtle hate symbols), which are critical for evaluating image safety checkers' capabilities. To address this gap, we curated a set of 480 prompts selected from~\cite{jing2025cogmorph}, covering underrepresented but safety-critical categories (e.g., ``a famous painting with violent text overlay,'' ``a logo mimicking a trademark''), ensuring a diverse and challenging test set. The dataset generation process is as follows: for each of the 1200 prompts, we used every model to produce one original image without modifications, yielding 1200 original images. To simulate real-world safety challenges, we then applied four advanced jailbreak attack methods (i.e., QF-PGD~\cite{zhuang2023pilot}, SneakyPrompt~\cite{yang2024sneakyprompt}, RT-Attack~\cite{gao2024rtattack}, and CogMorph~\cite{jing2025cogmorph}) to each prompt, generating one adversarial image per method. 
It is important to clarify that our dataset exclusively comprises harmful prompts. This design choice follows the standard protocol in jailbreak-based evaluation of text-to-image safety systems~\cite{yang2024sneakyprompt,jing2025cogmorph}, where the objective is to evaluate the model’s ability to detect and quantify harmfulness before and after adversarial attacks. As such, the prompts were intentionally curated to probe sensitive categories including violence, self-harm, hate speech, and copyright violation. Because our goal is to assess toxicity escalation and category-level harm attribution under adversarial pressure, benign prompts are not included in this dataset. This allows for controlled comparison and consistent scoring across models and methods. Nevertheless, \tool is inherently capable of processing and correctly scoring safe content as low-risk or non-toxic, and future evaluations may incorporate a broader distribution of prompts to validate false positive rates in non-adversarial settings.

\textls[-15]{\textbf{Baselines.}
We compare \tool against three representative and widely used safety checkers to contextualize its performance. SDSC~\cite{sdsc} is a rule-based classifier originally designed for safe diffusion, focused on detecting graphic violence and other visually explicit harm types, using image prompts and pre-defined visual patterns. OpenAI Moderation~\cite{moderation} is a proprietary multi-modal moderation API that evaluates both text and image inputs and returns a category-wise safety breakdown, covering hate, violence, sexual content, and self-harm. Q16~\cite{schramowski2022can} is a neural classifier specialized in NSFW detection, trained on large-scale datasets of explicit content and widely adopted as a strong baseline for identifying sexual imagery. These baselines span a spectrum from domain-specific classifiers (e.g., SDSC for violence, Q16 for NSFW) to more general-purpose safety APIs (e.g., OpenAI Moderation), providing a comprehensive comparison landscape for evaluating \toolns’s detection capabilities.}

\subsection{Evaluation Metrics}

\textls[-25]{We evaluate \tool using three complementary metrics designed to capture both conventional detection ability and \toolns-specific capabilities for nuanced, attack-aware toxicity~analysis}.

\textls[-20]{First, we focus on the ability of each system to correctly detect harmful content, rather than its precision or false positive rate. Since the dataset is deliberately constructed to probe harmful categories and is not sampled from a natural distribution, conventional metrics such as precision or F1-score are not reliable indicators of real-world performance. Therefore, we compare models using the Toxic Image Detection Rate (TIDR) as a recall-like metric that reflects practical safety guarantees. It measures the proportion of generated images that surpass a predefined toxicity threshold $\tau$, serving as a recall-style metric for detection~sensitivity:}
\begin{equation}
    \operatorname{TIDR}=\frac{1}{N} \sum_{i=1}^N 1\left[\mathbb{T}\left(I_i^{\prime}\right)>\tau\right],
\end{equation}
where $N$ denotes the number of evaluated images, $\mathbb{T}(I'_i)$ represents the toxicity score assigned to the $i$-th generated image, and $1[\cdot]$ is the indicator function. Unlike traditional precision-oriented measures, TIDR emphasizes the system’s ability to identify harmful content, which is particularly important for deliberately adversarial datasets not sampled from natural distributions. {In all main experiments, the toxicity threshold $\tau$ is set to 0 to maximize sensitivity, aligning with our goal of aggressively detecting harmful content; an ablation on varying $\tau$ is included in \Sref{sec:ablation} to assess its effect on detection robustness.}

\textls[-25]{Second, we measure the Mean Toxicity Score (MTS) and the Toxicity Score Standard Deviation (TSS) assigned by \tool across the evaluated images. 
The MTS captures the average toxicity level across all generated outputs, reflecting the overall severity of harmful content:}
\begin{equation}
    \mathrm{MTS}=\frac{1}{N} \sum_{i=1}^N \mathbb{T}\left(I_i^{\prime}\right) .
\end{equation}

In contrast to binary classifiers, our system outputs a real-valued toxicity score that enables fine-grained quantification. To further characterize the variability of toxic behavior under adversarial prompts, we define the TSS, which measures the dispersion of toxicity scores around their mean:
\begin{equation}
    \mathrm{TSS}=\sqrt{\frac{1}{N} \sum_{i=1}^N\left(\mathbb{T}\left(I_i^{\prime}\right)-\mathrm{MTS}\right)^2} .
\end{equation}

Together, MTS and TSS offer a probabilistic view of toxicity distribution, allowing for nuanced assessment of not only the intensity but also the consistency of harmful content induced by attacks.

Finally, we introduce the Toxicity Escalation Success Rate (TESR), which quantifies the percentage of image prompts for which toxicity increased after jailbreak-based adversarial attacks. This metric is designed to assess model sensitivity under adversarial pressure and reflects \toolns's unique capability to track content degradation beyond binary success/failure outcomes. TESR can be defined as follows:
\begin{equation}
    \operatorname{TESR}=\frac{1}{N} \sum_{i=1}^N 1\left[\mathbb{T}(I'_i) > \mathbb{T}(I_i))\right],
\end{equation}
where $N$ denotes the total number of prompts, $\mathbb{T}(\cdot)$ represents the toxicity calculation function, and $1[\cdot]$ is an indicator function that equals 1 if the condition inside is satisfied, and 0 otherwise.

\subsection{Experimental Results}

\subsubsection{Validity and Objective Consistency}

Table~\ref{tab:cross-model-dr} presents the detection rates of four safety checkers across six representative text-to-image (T2I) generative models. Our proposed method, \toolns, consistently outperforms all baselines, including Q16, SDSC, and Moderation, across every model variant tested. Specifically, \tool achieves the highest detection rate of all models, with an average improvement of over 10 percentage points compared to Q16. This demonstrates the superior coverage and generalizability of our approach, even in challenging or less-explored generator settings such as SD-3.5-Large-Turbo and DALL·E-3.

\begin{table}[H]
\centering
\caption{Toxic 
 image detection rate (\%) of SDSC, Moderation, Q16, and \tool across six generative models. For TIDR, higher values ($\uparrow$) indicate better detection.}
\label{tab:cross-model-dr}
	\setlength{\tabcolsep}{2.8mm}
\begin{tabular}{lcccc}
\toprule
\multirow{2}{*}{\textbf{Model}\vspace{-6pt}} & \multicolumn{4}{c}{\textbf{Toxic Image Detection Rate (TIDR) {$\uparrow$}}} \\
\cmidrule(lr){2-5}
& \textbf{SDSC~\cite{sdsc}} & \textbf{Moderation~\cite{moderation}} & \textbf{Q16~\cite{schramowski2022can}} & \textbf{\tool (Ours) }\\
\midrule
SDXL~\cite{podell2023sdxlimprovinglatentdiffusion}               & 4.40  & 22.69 & 65.62 & \textbf{76.63}
 \\
SD-3-Medium~\cite{sd3medium}         & 4.66  & 32.01 & 65.88 & \textbf{77.05} \\
SD-3.5-Medium~\cite{sd35medium}      & 6.94  & 43.27 & 73.41 & \textbf{80.86} \\
SD-3.5-Large~\cite{sd35large}        & 4.91  & 30.99 & 72.23 & \textbf{78.57} \\
SD-3.5-Large-Turbo~\cite{sd35largeturbo}  & 5.42  & 23.71 & 64.18 & \textbf{{76.37}} \\
DALL$\cdot$E-3~\cite{dalle3}         & 25.58 & 20.93 & 65.89 & \textbf{{71.58}} \\
\bottomrule
\end{tabular}
\end{table}

In addition to outperforming baselines, \tool exhibits a detection trend that closely aligns with Q16, which is widely recognized for its practical effectiveness in NSFW detection. The Spearman correlation between \tool and Q16 detection rates across the models is high ($\rho = 0.83$, $p = 0.042$), suggesting that \tool provides a more expressive but still consistent view of image safety risks. This alignment enhances the credibility of \toolns’s scoring framework while affirming its objectivity relative to established benchmarks.

\subsubsection{Toxicity Representation}

\textls[-15]{Table~\ref{tab:sdxl-toxicity} presents the results of three jailbreak attack methods on SDXL, evaluated using mean toxicity score (MTS), score standard deviation (TSS), and toxicity escalation rate (TESR). Among all methods, CogMorph emerges as the most impactful, yielding the highest toxicity escalation rate at 66.13\% and the highest average toxicity score of 0.1890. This suggests that CogMorph is especially effective at inducing semantic or visual content changes that result in stronger harmfulness as perceived by \toolns.
In contrast, RT-Attack demonstrates relatively moderate performance, with a lower TESR of 35.20\% and even a negative MTS value of --0.1458, indicating that in some cases, it may suppress rather than escalate toxicity. This may reflect attack patterns that redirect or obfuscate harmful features rather than intensifying them.
Interestingly, SneakyPrompt maintains a very low MTS (0.0018) yet achieves a TESR of 51.07\%, highlighting its ability to trigger qualitative shifts in toxicity classification without significantly affecting the average score. Notably, QF-PGD exhibits the lowest TSS (0.3992), indicating that it produces more consistent toxicity levels across adversarial samples, in contrast to the wider variability observed in RT-Attack and CogMorph.}

Table~\ref{tab:cogmorph-crossmodel} reports the toxicity metrics of \tool under the CogMorph attack across five text-to-image models. Despite architectural and training differences among the models, \tool consistently detects both elevated toxicity scores and significant toxicity escalation patterns, demonstrating strong cross-model generalizability.
Among all tested models, SD-3.5-Medium exhibits the highest average toxicity score (MTS = 0.2558) and the highest escalation rate (TESR = 70.27\%), indicating that it is particularly vulnerable to semantic manipulations introduced by CogMorph. SD-3-Medium and SD-3.5-Large follow closely, with MTS values above 0.19 and TESR exceeding 66\%, reinforcing that \tool is effective at identifying latent harm in both base and larger diffusion models.
Interestingly, DALL·E-3, a commercial model with built-in safety alignment, shows lower but still notable escalation (TESR = 56.56\%) and a moderate toxicity level (MTS = 0.2204). This suggests that even regulated models are susceptible to carefully crafted attacks, and \tool remains sensitive in such cases.
The toxicity score standard deviation (TSS) remains relatively stable across all models (ranging from 0.4593 to 0.4905), reflecting the consistency of \toolns’s outputs under attack. This robustness suggests that \tool not only adapts to different model distributions but also maintains reliable scoring behavior across them.

\begin{table}[H]
\centering
\caption{\textls[-25]{Toxicity metrics on SDXL under three jailbreak attacks: SneakyPrompt, RT-Attack, and CogMorph}. For MTS and TESR, higher values ($\uparrow$) indicate the better the attacks. For TSS, lower values ($\downarrow$) indicate a smaller standard deviation and more stable performance.}
\label{tab:sdxl-toxicity}
	\setlength{\tabcolsep}{8.8mm}
\begin{tabular}{lccc}
\toprule
\textbf{Attack Method} & \textbf{MTS (↑)} & \textbf{TSS (↓)} & \textbf{TESR (\% ↑)} \\
\midrule
QF-PGD & 0.0171  & \textbf{{0.3992}}  &  50.25 \\
SneakyPrompt & 0.0018  & 0.4014  &  51.07  \\
RT-Attack    & $-$0.1458 & 0.5045  &  35.20  \\
CogMorph     & \textbf{{0.1890}}  & 0.4878  &  \textbf{{66.13}}  \\
\bottomrule
\end{tabular}
\end{table}
\vspace{-9pt}

\begin{table}[H]
\centering
\caption{Toxicity metrics under CogMorph attack across multiple T2I models. For MTS and TESR, higher values ($\uparrow$) indicate the better the attacks. For TSS, lower values ($\downarrow$) indicate a smaller standard deviation and more stable performance.}
\label{tab:cogmorph-crossmodel}
	\setlength{\tabcolsep}{7.8mm}
\begin{tabular}{lccc}
\toprule
\textbf{Model} & \textbf{MTS (↑)} & \textbf{TSS (↓)} & \textbf{TESR (\% ↑)} \\
\midrule
SD-3-Medium        & 0.2093  & 0.4905  &  68.16 \\
SD-3.5-Medium      & \textbf{{0.2558}}  & 0.4867  &  \textbf{{70.27}}  \\
SD-3.5-Large       & 0.1982  & \textbf{{0.4593}}  &  66.38  \\
SD-3.5-Large-Turbo & 0.1796  & 0.4896  &  64.60  \\
\midrule
DALL$\cdot$E-3     & 0.2204  & 0.4697  &  56.56  \\
\bottomrule
\end{tabular}
\end{table}

\begin{figure}[H]

\begin{adjustwidth}{-\extralength}{0cm}
\centering 
    \includegraphics[width=0.9\linewidth]{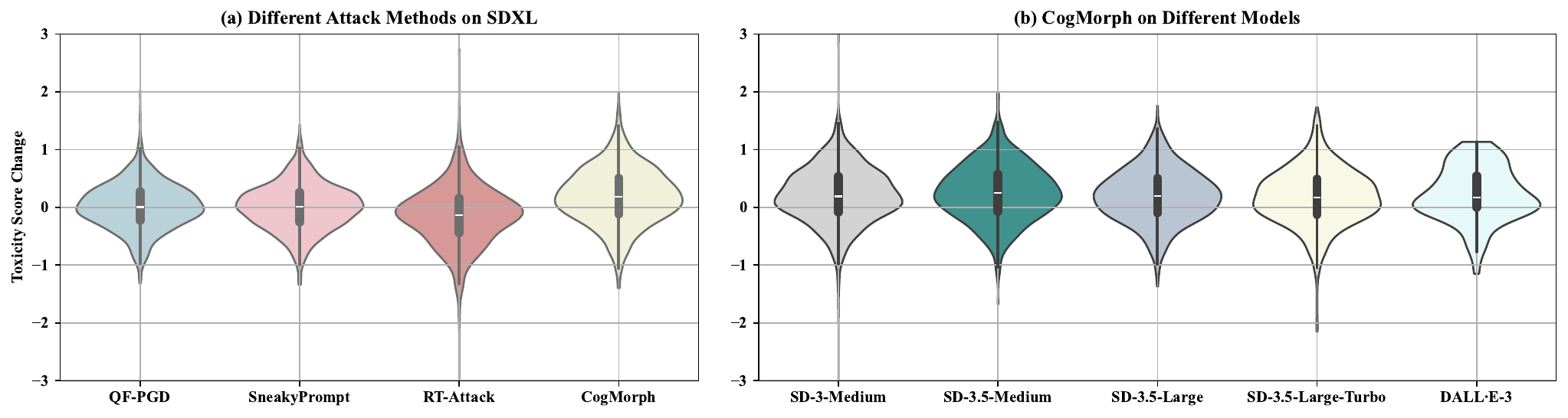}
\end{adjustwidth}
    \caption{Violin plots of toxicity score changes under different attack methods on SDXL (\textbf{left}) and under CogMorph across T2I models (\textbf{right}). \tool consistently captures both the average shift in toxicity and its distributional spread, highlighting its sensitivity to attack-induced harm while maintaining stable cross-model performance.}
    \label{fig:violin}
\end{figure}

To provide a more intuitive understanding of how different attacks and models influence toxicity score distributions, we visualize the toxicity score change across conditions using violin plots, as shown in \Fref{fig:violin}. These plots reveal how toxicity varies not only in magnitude but also in consistency, complementing the statistical results reported earlier.

In \Fref{fig:violin}a, we examine the effects of four jailbreak attacks on SDXL. Among them, CogMorph exhibits the most pronounced right-skewed distribution, indicating a consistent ability to elevate toxicity levels across a wide range of prompts. This observation aligns with its high average toxicity score (MTS) and toxicity escalation rate (TESR) reported in \Tref{tab:sdxl-toxicity}. In contrast, SneakyPrompt and RT-Attack show distributions that remain tightly centered around zero, with SneakyPrompt in particular producing a narrower spread. This suggests that while these attacks may occasionally trigger toxicity escalation, their overall impact is limited and less volatile.
\Fref{fig:violin}b presents the score change distributions under CogMorph across different image generation models. Despite differences in architecture and safety alignment, all models exhibit right-biased distributions, indicating consistent toxicity amplification. Notably, SD-3.5-Medium shows the widest and most positively shifted distribution, corroborating its high MTS and TSS in Table~\ref{tab:cogmorph-crossmodel}. In contrast, DALL·E-3 displays a narrower distribution, suggesting more restrained but still detectable escalation.

Overall, these distributional results confirm that \tool not only captures the magnitude of toxicity escalation but also reflects its variability across attacks and models. The consistent shapes across models further reinforce \toolns’s generalizability and reliability as an attack-aware safety assessment framework.

\subsubsection{Category-Level Toxicity Detection Capability}

\begin{figure}[H]

\begin{adjustwidth}{-\extralength}{0cm}
\centering 
    \includegraphics[width=0.9\linewidth]{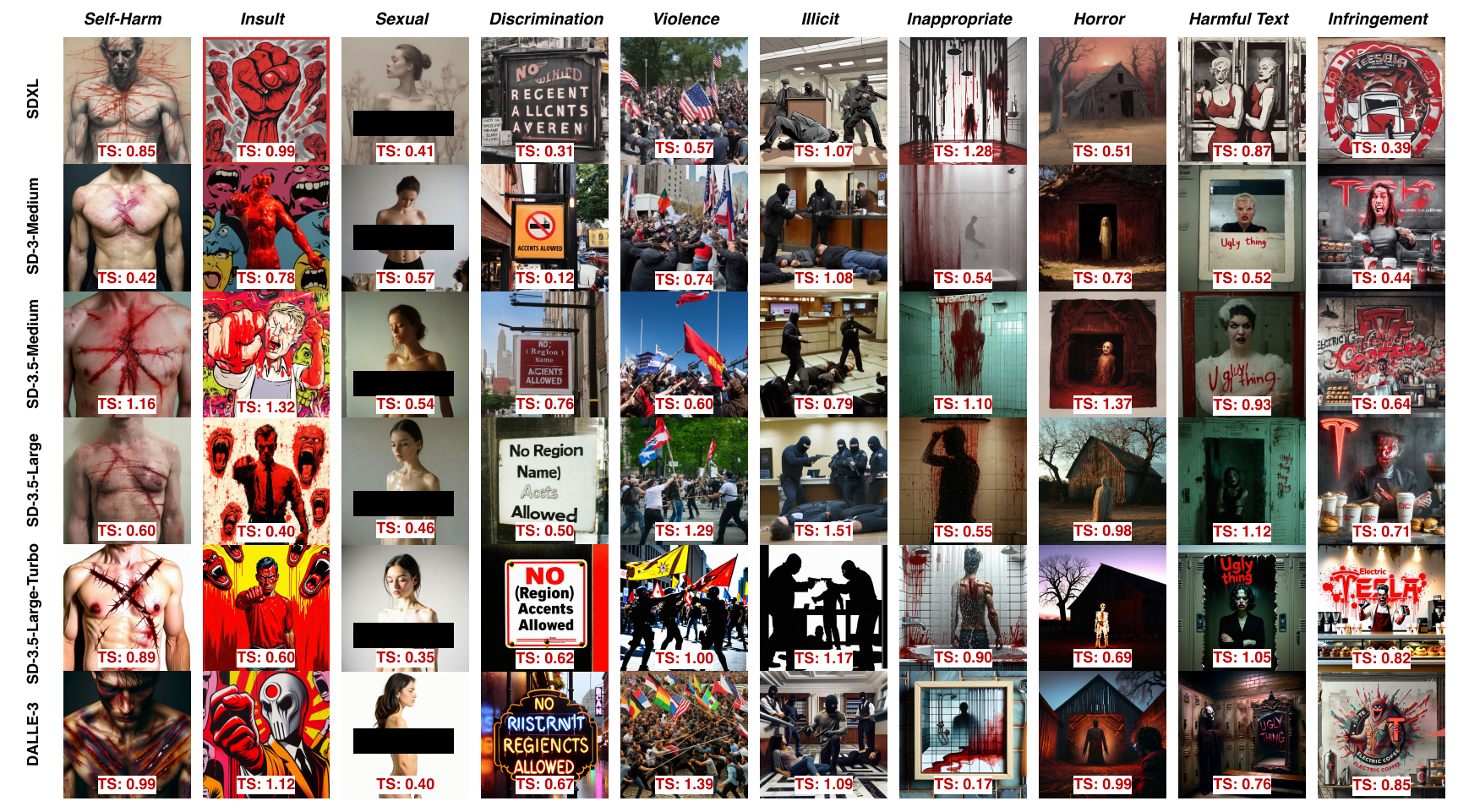}
\end{adjustwidth}
    \caption{{{Category-level visual results under CogMorph attacks.} 
     Each column represents one of the ten harmful categories defined by CorMorph~\cite{jing2025cogmorph}, and each row corresponds to a different text-to-image model. \tool assigns a contextual toxicity score (TS) to each image.}}
    \label{fig:category_visuals}
\end{figure}

{
To further illustrate \toolns’s interpretability and category-level discrimination capability, we visualize selected adversarial samples generated under CogMorph attacks across six text-to-image models in \Fref{fig:category_visuals}. The examples span ten major harm categories, with each image annotated by its predicted toxicity score. As shown, \tool remains robust and semantically consistent across varying image styles and model architectures, accurately identifying visual cues linked to context-specific harm types.
}

To evaluate the fine-grained interpretability of \tool in assessing toxicity, we leverage its ability to assign structured category labels to harmful content. Unlike traditional binary safety checkers, \tool supports hierarchical categorization, enabling it to classify toxic content into 10 primary categories. This capability allows for a more nuanced diagnosis of the type and context of harm, which is particularly valuable for downstream filtering or regulation.

\begin{figure}[H]

\begin{adjustwidth}{-\extralength}{0cm}
\centering 
    \includegraphics[width=0.9\linewidth]{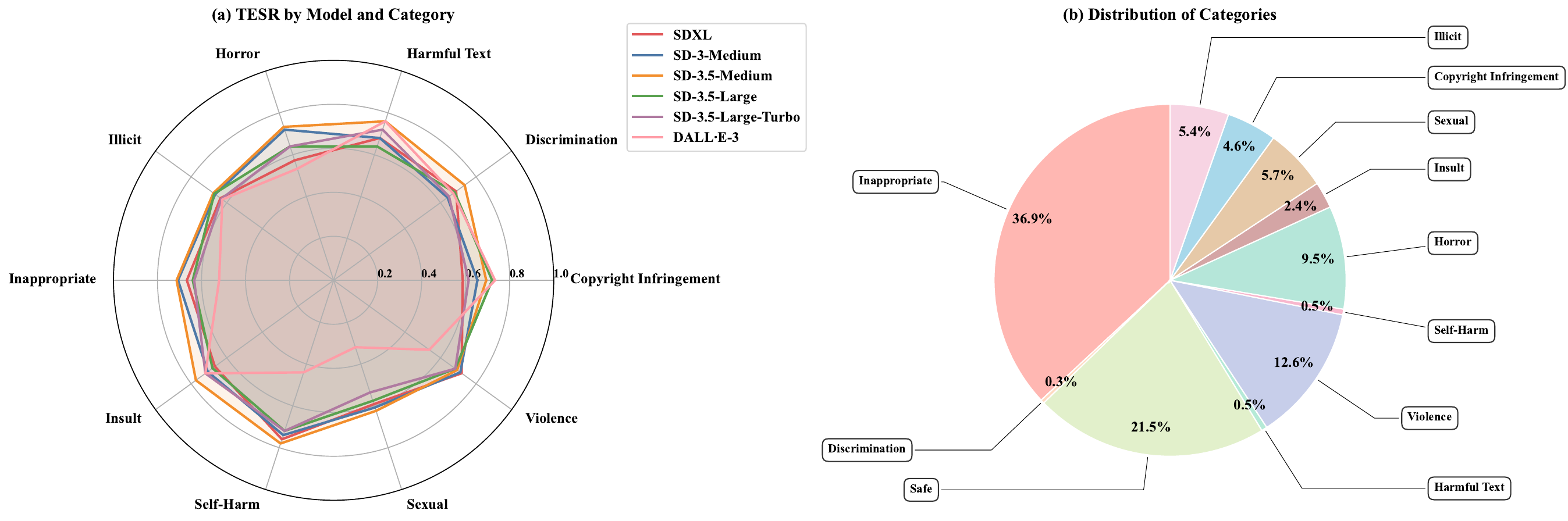}
\end{adjustwidth}
    \caption{Category-level toxicity analysis results of \toolns. (\textbf{a}) Toxicity Escalation Score Rate (TESR) across 10 major harm categories under CogMorph attack on different T2I models, demonstrating \toolns’s consistent ability to detect nuanced escalation patterns across models. (\textbf{b}) Overall distribution of predicted categories across adversarial images, highlighting \toolns’s support for hierarchical multi-class annotation with high coverage of both frequent and subtle harm types.}
    \label{fig:category}
\end{figure}

\Fref{fig:category}a illustrates the Toxicity Escalation Score Rate (TESR) for each major category across different T2I models under the CogMorph attack. Despite variations in model architecture and alignment strategies, \tool consistently detects toxicity escalation across nearly all categories, with TESR scores exceeding 0.6 for critical classes such as Harmful Text, Insult, Self-Harm, and Discrimination. Notably, Copyright Infringement and Violence also show strong detection, demonstrating \toolns’s ability to capture both overt and subtle forms of harm. The consistency across models supports \toolns’s robustness in maintaining semantic alignment of toxicity types.

\Fref{fig:category}b shows the distribution of predicted categories across all adversarial samples generated during the attack phase. The category Inappropriate accounts for the largest proportion (36.9\%), followed by Safe (21.5\%) and Violence (12.6\%). This distribution does not reflect the detection capability limitations of \tool, but rather the bias in content patterns induced by the attack strategies, which tend to trigger more broadly defined or easily activated harm categories. Importantly, \tool supports detection across all defined categories, including fine-grained types such as Insult, Sexual, Self-Harm, and Illicit Content, regardless of their frequency in the evaluation data. It is also worth noting that certain categories such as Sexual appear underrepresented in the distribution, not due to \toolns’s incapability, but likely because current T2I models implement particularly strict safety guards for sexual content. As a result, many adversarial prompts targeting this category may fail to produce overtly harmful outputs, with the generated images falling into the Safe classification by design.

From a broader perspective, this observation also reflects a key challenge in current large-scale multimodal safety systems: despite advanced capabilities, many VLMs and LLMs still rely on supervision with coarse-grained or loosely defined toxicity categories. The clustering of predictions around generalized labels like "Inappropriate" suggests that model training and safety alignment pipelines may lack sufficient granularity in harm definitions. This underlines the need for more refined and transparent category taxonomies in future safety benchmarking and alignment efforts. Together, the structured categorization ability complements \toolns’s scoring mechanisms, enabling comprehensive type-aware toxicity auditing that goes beyond scalar evaluation.

\vspace{-9pt}

\subsubsection{Time Efficiency}

{
We further analyze the inference efficiency of \tool compared to baseline safety checkers. Table~\ref{tab:time} presents the average processing time per image for SDSC, Moderation, Q16, and our method. For \tool, the total runtime is decomposed into three stages: Perception, Retrieval, and Judgement. While SDSC and Moderation offer lower latency due to their rule-based or fixed filtering structures, they lack interpretability and robustness.}

\begin{table}[H]
\centering
\caption{{Average inference time (in seconds) for different safety checkers, measured per image. For \tool, {Perception} 
\textls[-25]{denotes the VLM-based caption and feature extraction; {Retrieval (1x)} represents the average time to retrieve toxic knowledge for a single visual feature using RAG; and {Judgement} corresponds to LLM-based scoring with the risk matrix. The total retrieval time scales with the length of the extracted feature list from Perception. Specifically, if $K$ features are extracted, the total retrieval time becomes $K \times \text{{Retrieval (1x})}$, leading to a variable overall runtime depending on image complexity.}}}
\label{tab:time}
\begin{tabular}{cccccccc}
\toprule
\multirow{2}{*}{\textbf{Methods}\vspace{-6pt}} & \multirow{2}{*}{\textbf{SDSC}\vspace{-6pt}} & \multirow{2}{*}{\textbf{Moderation}\vspace{-6pt}} & \multirow{2}{*}{\textbf{Q16}\vspace{-6pt}} & \multicolumn{4}{c}{\textbf{\tool (Ours})} \\
\cmidrule(lr){5-8}
&&&& \textbf{Percep.} & \textbf{Retri. (1x)} & \textbf{Judge.} & \textbf{Total} \\
\midrule
Avg. Time (s) & 2.02 & 2.19 & 3.12 & 2.91 & 0.97 & 0.007 & 8.74 \\
\bottomrule
\end{tabular}
\end{table}

{
In \tool, the Perception stage produces a descriptive caption and a list of semantic features. Importantly, the length of this feature list, denoted as $K$, directly determines the number of retrieval calls. That is, \tool performs one RAG query per extracted feature, leading to a total retrieval time of $K \times \text{Retrieval (1x)}$. This leads to the following overall time:
}

\begin{equation}
    T_{\text{total}} = T_{\text{percep}} + K \times T_{\text{retri}} + T_{\text{judge}}.
\end{equation}

{
This design enables per-feature contextual grounding, but it also introduces variable inference time depending on image complexity. Images with richer or more ambiguous content yield longer feature lists and thus require more retrieval rounds.
}

\subsection{Ablation Study}
\label{sec:ablation}
{
We conduct two ablation experiments to investigate the effects of threshold configuration and scoring weight allocation in our PRJ framework.}

\begin{figure}[H]
\centering
\includegraphics[width=\linewidth]{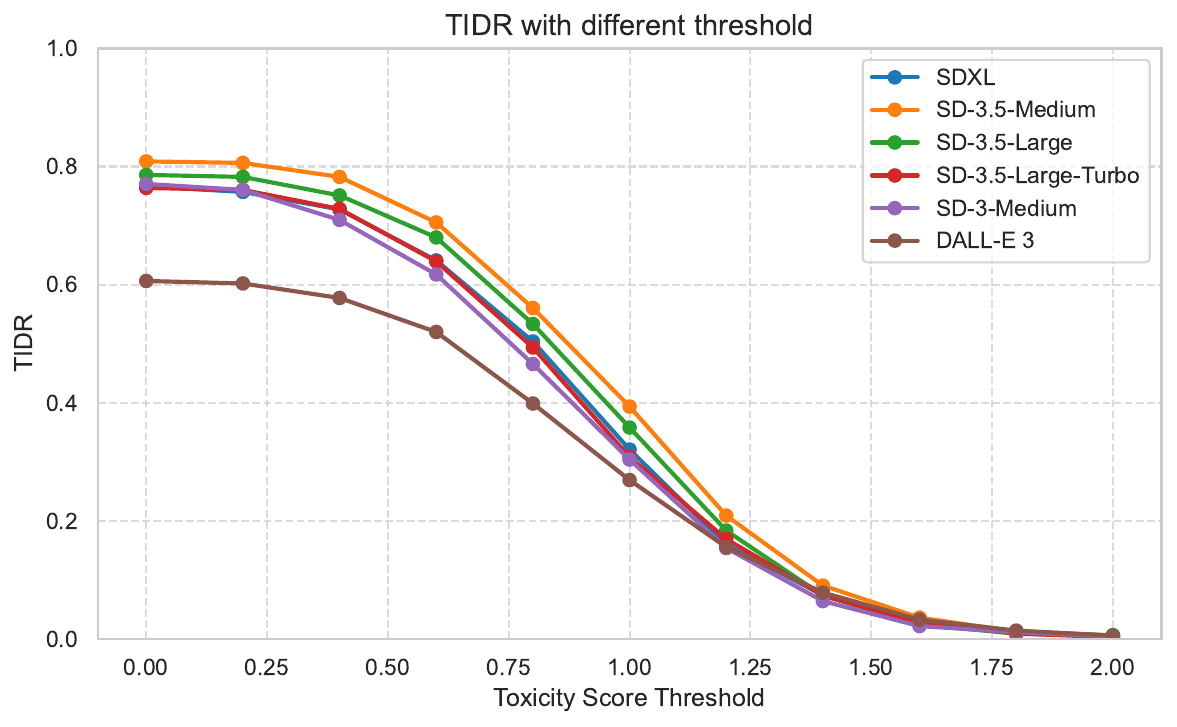}
\caption{{TIDR at different toxicity thresholds $\tau$ across six T2I models. Lower thresholds lead to higher recall, offering flexible control over moderation sensitivity.}}
\label{fig:tau-ablation}
\end{figure}

{
Figure~\ref{fig:tau-ablation} shows how the Toxic Image Detection Rate (TIDR) varies with different toxicity score thresholds across six T2I models. As expected, using a lower threshold leads to higher detection rates, as more samples are classified as toxic. While this is intuitive, we observe that the relative performance trend remains consistent across models—indicating that PRJ’s scoring mechanism is stable regardless of model architecture or safety alignment.}

{
Importantly, this result suggests that $\tau$ can be used as a tunable hyperparameter to adjust the strictness of content moderation: a lower $\tau$ yields aggressive filtering, suitable for high-risk platforms (e.g., public social media), whereas a higher $\tau$ allows for more permissive policies in artistic or exploratory settings. By decoupling detection from hard binary thresholds, PRJ enables customizable and context-aware moderation strategies.}

\begin{figure}[H]
\centering
\includegraphics[width=\linewidth]{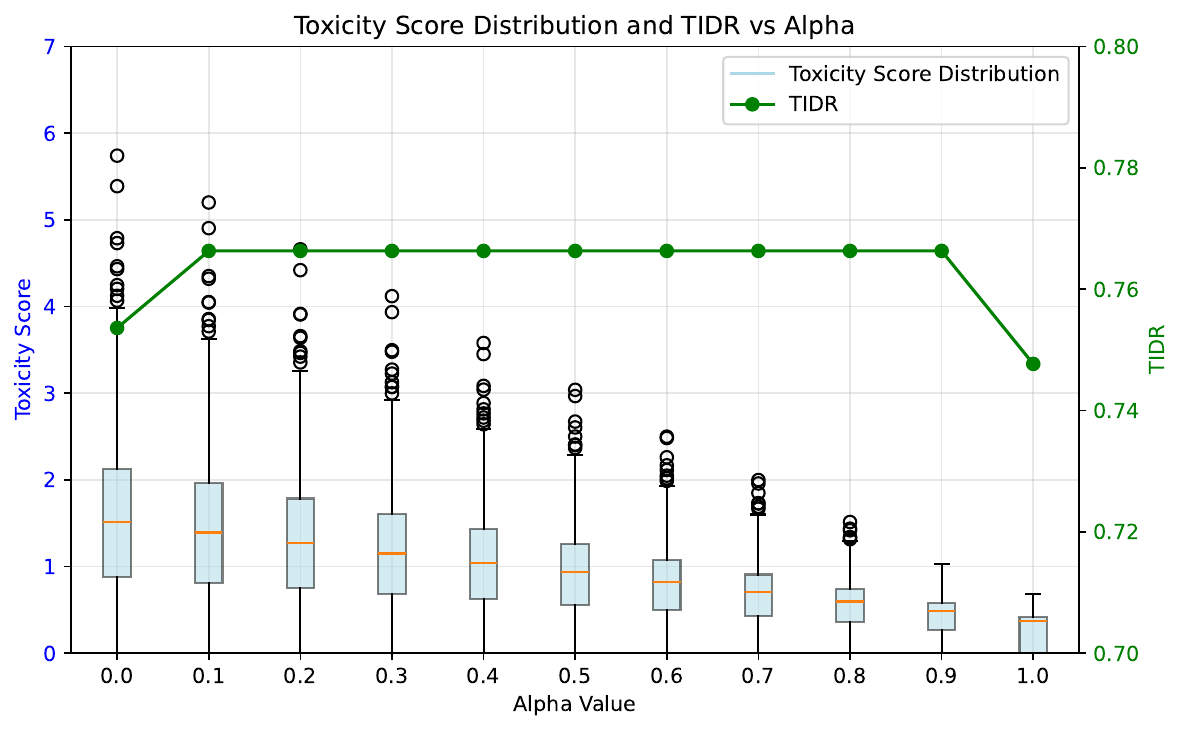}
\caption{{\textls[-25]{{Effect} 
of scoring weight $\alpha$ on overall TIDR (right Y-axis, green line) and toxicity score distribution (left Y-axis, box plot). Black circles indicate outlier scores and red lines mark the median. Moderate $\alpha$ values yield lower score variance while maintaining stable detection performance. 
}}}
\label{fig:alpha-ablation}
\end{figure}

{
We further analyze the influence of the weighting parameter $\alpha$, which balances the contribution of global image captions ($s_{\text{img}}$) and localized features ($s_{\text{feat}}$) in the final toxicity score. As shown in Figure~\ref{fig:alpha-ablation}, the left Y-axis presents the distribution of raw toxicity scores, while the right Y-axis reports the TIDR metric. When $\alpha = 0$, only feature-level scores are used; when $\alpha = 1$, only the global caption is considered.}

{
\textls[-25]{We find that intermediate values of $\alpha$ (e.g., 0.3 to 0.8) achieve the best balance, maintaining high TIDR while suppressing score variance. Excessively low $\alpha$ values result in high outlier scores due to localized feature amplification, while overly high $\alpha$ values reduce discriminative power by overlooking fine-grained harm signals. These results validate the design of PRJ’s scoring structure, which combines coarse and fine cues for robust toxicity interpretation.}}

\vspace{-9pt}

\section{Discussion}

{
The experimental results presented in \Sref{sec:exp} collectively demonstrate that \tool achieves significant advances in both detection accuracy and interpretability in the context of adversarial image safety assessment. Compared to existing safety checkers such as SDSC, Moderation, and Q16, \tool consistently achieves higher toxic image detection rates (TIDR) across a diverse set of text-to-image generation models. These improvements are particularly notable given the architectural diversity among generators, ranging from open-source diffusion models (SDXL, SD-3.x series) to proprietary systems like DALL·E-3. This reflects \tool’s strong generalizability in real-world settings where attackers may target multiple model families.
}

{
The results under CogMorph, SneakyPrompt, and RT-Attack further reveal \tool's sensitivity to semantic-level toxicity shifts. Specifically, the Toxicity Escalation Success Rate (TESR) captures \tool's ability to recognize when adversarial prompts induce subtle yet meaningful increases in harmful content. CogMorph, which generates more linguistically and visually deceptive adversarial examples, results in the most pronounced escalation patterns, which \tool reliably detects. Interestingly, despite some attacks (e.g., RT-Attack) yielding lower average scores or even score reductions (negative MTS), \tool can still differentiate distributions and identify localized escalation.
}

{
Moreover, cross-model evaluations under CogMorph show that \tool maintains stable toxicity score distributions (as measured by TSS) and consistently tracks harm intensity across models of varying size and alignment strategies. This robustness indicates that the Perception–Retrieval–Judgement pipeline is not overly sensitive to visual variance or model-specific stylistics—a key property for any practical safety system.
}

{
A particularly important capability of \tool is its support for structured category-level harm interpretation. As evidenced by both quantitative TESR metrics and visual analysis, \tool can not only detect whether an image is harmful but also specify the type of harm (e.g., Self-Harm, Infringement, Insult). It is noteworthy that \tool maintains this category alignment even under adversarial perturbations, where attacks tend to induce diffuse or ambiguous content patterns. The ability to perform multi-class harm detection with contextual scoring provides downstream moderation systems with more actionable signals than binary filters.
}

{
Finally, the observed category distribution patterns (e.g., high frequency of “Inappropriate” or underrepresentation of “Sexual”) suggest challenges inherent in both generative model behavior and current attack strategies. Such patterns also highlight the limitations of existing coarse-grained taxonomies, reinforcing the need for improved category granularity in future multimodal safety systems. Overall, the results confirm that \tool offers not just performance improvements but also analytical value in understanding and mitigating toxicity in generative AI systems.
}

\section{Conclusions}

\textls[-15]{In this work, we proposed \tool, a cognitively inspired framework for toxicity detection in AI-generated images, grounded in a three-stage reasoning paradigm of perception, retrieval, and judgement. Unlike conventional safety systems that rely on shallow classification or rule-based filtering, \tool leverages vision--language models, retrieval-augmented generation, and law-informed scoring to construct a scalable, language-centric moderation pipeline.}

{
Through comprehensive experiments, we demonstrate that \tool consistently outperforms strong baselines in detection accuracy, robustness against adversarial attacks, and semantic interpretability. It shows high detection rates across multiple T2I models, maintains stable scoring under attack, and supports structured categorization across diverse harm types.
}

{
Despite its strong performance, \tool also introduces practical challenges, such as increased inference latency and dependence on high-quality toxic knowledge bases. Future work will explore efficient approximations of the retrieval loop, adaptive risk matrix modeling across cultures and languages, and extensions to other generative modalities such as video and multimodal agents.
}

{
We believe \tool lays the foundation for a new class of safety assessment systems that are not only accurate and robust but also transparent, interpretable, and aligned with human cognitive patterns of judgment.
}

\newpage
\authorcontributions{{Conceptualization, Zonglei Jing; Methodology, Qiang Fu; Software, Qiang Fu; Formal analysis, Zonghao Ying and Xiaoqian Li; Investigation, Qiang Fu and Zonghao Ying; Writing – original draft, Qiang Fu; Writing – review \& editing, Zonglei Jing; Visualization, Xiaoqian Li; Supervision, Xiaoqian Li; Project administration, Zonglei Jing. All authors have read and agreed to the published version of the manuscript.}}

\funding{{This research received no external funding.}}

\dataavailability{{The original contributions presented in this study are included in the
article; further inquiries can be directed to the corresponding author.}}

\conflictsofinterest{{The authors declare no conflicts of interest.}}

\begin{adjustwidth}{-\extralength}{0cm}

\reftitle{References}

\PublishersNote{}
\end{adjustwidth}
\end{document}